\pdfoutput=1

\documentclass[11pt]{article}

\usepackage[]{emnlp2021}

\usepackage{times}
\usepackage{latexsym}

\usepackage[T1]{fontenc}

\usepackage[utf8]{inputenc}

\usepackage{microtype}

\usepackage{microtype}
\usepackage{booktabs}
\usepackage{multirow}
\usepackage{graphicx}
\usepackage{caption}
\usepackage[normalem]{ulem}
\usepackage[utf8]{inputenc}
\usepackage{dingbat}
\usepackage{wrapfig}
\usepackage{amssymb}
\usepackage{amsfonts}
\usepackage{amsmath}
\usepackage{amssymb}
\usepackage{amsthm}
\usepackage{mathrsfs}
\usepackage{adjustbox}
\usepackage{bbold}
\usepackage[font=small]{caption}
\usepackage[linesnumbered,ruled]{algorithm2e}
\usepackage{url}
\usepackage{subcaption}
\usepackage{xspace}

\title{Efficient Test Time Adapter Ensembling \\ for Low-resource Language Varieties}

\author{
\textbf{Xinyi Wang}$^{1}$ \quad

\textbf{Yulia Tsvetkov}$^{2}$ \quad

\textbf{Sebastian Ruder}$^{3}$ \quad

\textbf{Graham Neubig}$^{1}$
\\
$^{1}$Language Technology Institute, Carnegie Mellon University \\
$^{2}$Paul G.~Allen School of Computer Science \& Engineering, University of Washington \\
$^{3}$DeepMind \\
\texttt{xinyiw1@cs.cmu.edu,yuliats@cs.washington.edu} \\
\texttt{ruder@google.com,gneubig@cs.cmu.edu}
}

\begin{document}
\maketitle
\begin{abstract}
     Adapters are light-weight modules that allow parameter-efficient fine-tuning of pretrained models. Specialized language and task adapters have recently been proposed to facilitate cross-lingual transfer of multilingual pretrained models~\citep{madx-pfeiffer-etal-2020}. However, this approach requires training a separate language adapter for every language one wishes to support, which can be impractical for languages with limited data. An intuitive solution is to use a related language adapter for the new language variety, but we observe that this solution can lead to sub-optimal performance. In this paper, we aim to improve the robustness of language adapters to uncovered languages without training new adapters. We find that ensembling multiple existing language adapters makes the fine-tuned model significantly more robust to other language varieties not included in these adapters.
     Building upon this observation, we propose Entropy Minimized Ensemble of Adapters (EMEA), a method that optimizes the ensemble weights of the pretrained language adapters for each test sentence by minimizing the entropy of its predictions. Experiments on three diverse groups of language varieties show that our method leads to significant improvements on both named entity recognition and part-of-speech tagging across all languages.    
\end{abstract}
\section{Introduction}
Massively multilingual pretrained models~\citep{bert,uniencoder,xlm,xlmr} combined with cross-lingual transfer now define the state of the art on a variety of NLP tasks~\citep{xtreme2020}. 
Within this paradigm, multilingual pretrained models are fine-tuned on annotated data of a task in a high-resource language, and transferred to other languages.
Several recent works propose parameter-efficient fine-tuning methods that insert small \emph{adapter} modules between the layers of pretrained models~\citep{adapter-vision,adapter-houlsby}. In this line of work, the pretrained model is usually frozen while only the adapters are fine-tuned for a downstream task, 
which is conducive to both improving the model's learning ability and compactness with respect to storage on disk or in memory. The adapters can be applied to the cross-lingual transfer setting by training separate language and task adapters~\cite{madx-pfeiffer-etal-2020,udapter}. 
Specifically, \citet{madx-pfeiffer-etal-2020} propose to perform zero-shot transfer by first training language-level adapters on monolingual data in different languages and then a task adapter on annotated data in the source language. 

\begin{figure}
    \centering
    \includegraphics[width=0.35\textwidth]{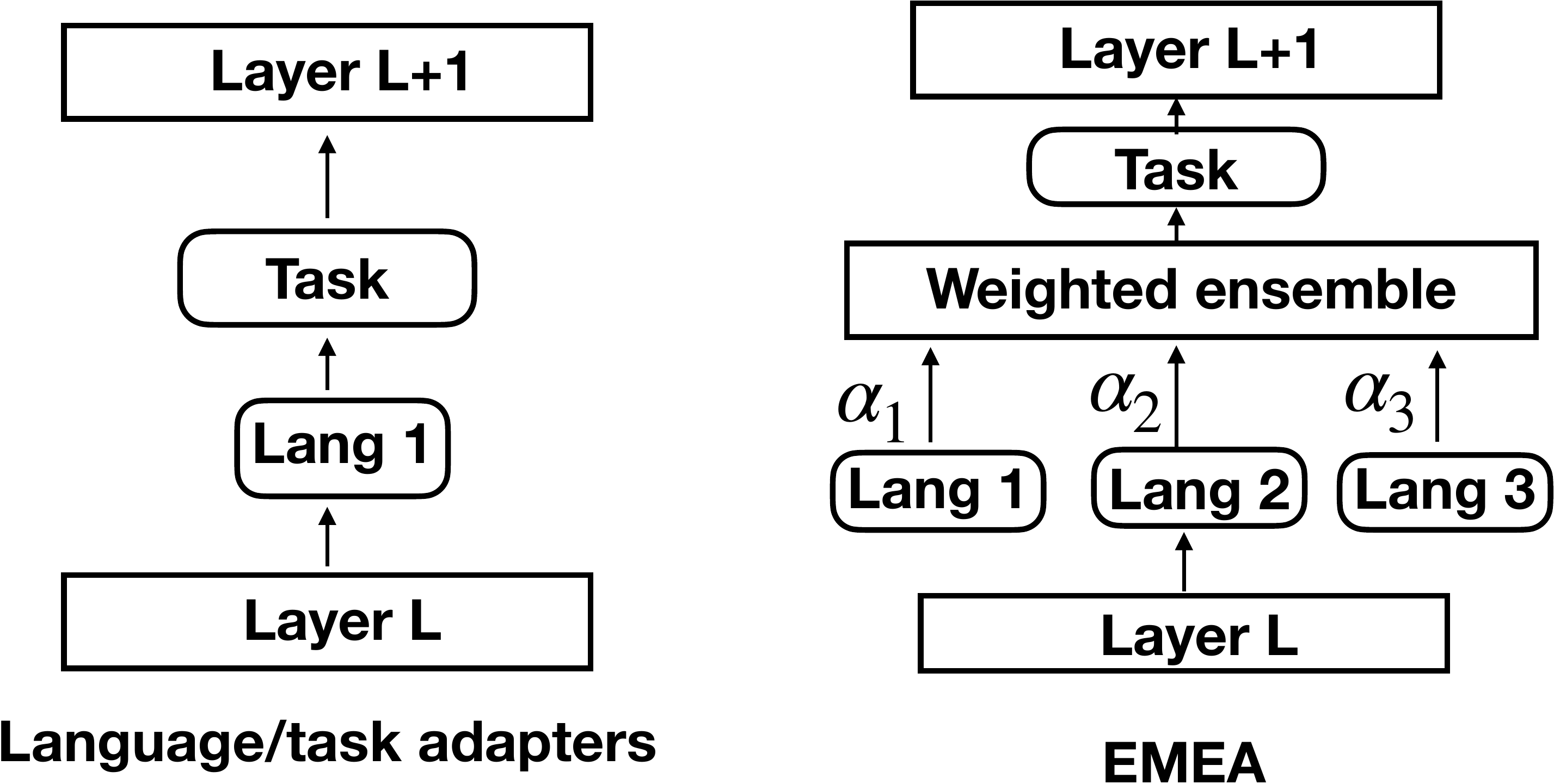}
    \caption{Comparison of the standard cross-lingual adapter and our method of entropy minimized ensembling of adapters (EMEA), which combines multiple language adapters to improve robustness to new language varieties at \textit{test time}.}
    \label{fig:method}
    \vspace{-2em}
\end{figure}

One drawback of this framework is that a separate language adapter is required for each target language, which is  problematic in cases where the data to train these adapters cannot be easily obtained, such as for languages with diverse regional or demographic variations. In fact, certain language varieties are not included in the standard language identification tools, which makes it challenging to reliably obtain even unlabeled data~\citep{salameh-etal-2018-fine,language_id_wild,learn_dialect_feature}. 
To give just one example, the Nordic languages and dialects form a dialect continuum where the total number of language varieties is difficult to estimate, and language varieties constantly emerge in culturally and linguistically diverse areas~\citep{norway_multilingual,norway_dialect_acquire}. 
Although highly related, these language varieties have many systematic differences, which need to be addressed by NLP systems that equitably serve all speakers \citep{kumar2021langvar}. 
One potential mitigation strategy is directly using an adapter trained on another similar language variety, but we find this sub-optimal in experiments (\autoref{sec:exp}). 

Instead, we propose two methods to combine existing language adapters to \emph{adapt the model to new language varieties at test time} without any training data.
First, we find that simply ensembling multiple related language adapters can significantly improve the fine-tuned model, compared with using individual language adapters.
Second, we propose Entropy Minimized Ensemble of Adapters~(EMEA; \autoref{fig:method}), which adapts the ensemble weight of the language adapters for each test instance by minimizing the ensembled model's prediction uncertainty.
Our experiments show that EMEA further improves over vanilla ensembling for three groups of uncovered language varieties on both the named entity recognition and part-of-speech tagging tasks.

\section{Adapters for Cross-lingual Transfer}
To facilitate our discussion, we briefly summarize the MAD-X framework~\citep{madx-pfeiffer-etal-2020} for zero-shot cross-lingual transfer and identify its shortcomings. 
The goal of MAD-X is to fine-tune a multilingual pretrained model $\mathcal{M}$ to $m$ downstream tasks $T_1,T_2,...,T_m$, each of which could be in $n$ languages $L_1,L_2,...,L_n$. To this end, MAD-X relies on language and task adapters, which are light-weight functions inserted in the Transformer layers in $\mathcal{M}$---usually a feed-forward down-projection followed by an up-projection.
Specifically, let $h$ be the output of an intermediate layer in $\mathcal{M}$, then $\mathcal{L}_j(h)$ is the transformation that projects $h$ into the embedding space for language $L_j$, and $\mathcal{T}_i(\mathcal{L}_j(h))$ is the transformation that projects $\mathcal{L}_j(h)$ into the embedding space for task $T_i$. 

MAD-X trains the adapters $\mathcal{T}_i(\cdot)$ and $\mathcal{L}_j(\cdot)$ in two steps. First, for each language $L_j$, its adapter $\mathcal{L}_j$ is inserted into $\mathcal{M}$ to replace the output of each layer $h$ with $\mathcal{L}_j(h)$. The resulting model, which we denote as $\mathcal{L}_j \circ \mathcal{M}$, is trained on unlabeled data in $L_j$ using an unsupervised objective such as masked language modeling \cite[MLM;][]{bert}. Second, for each task $T_i$, its adapter $\mathcal{T}_i$ is inserted on top of a \textit{src} language adapter $\mathcal{L}_\text{src}$. The resulting model $\mathcal{T}_i \circ \mathcal{L}_\text{src} \circ \mathcal{M}$ is trained on the downstream task $T_i$ in language $L_\text{src}$. 
After these two steps, $\mathcal{T}_i \circ \mathcal{L}_j \circ \mathcal{M}$ can be used to perform zero-shot cross-lingual transfer for any task $T_i$ and language $L_j$.

\paragraph{Shortcomings} 
This approach requires a separate adapter for each language one wishes to support. 
The online database AdapterHub\footnote{\url{https://adapterhub.ml/}} aims to improve the efficiency and reuse of trained language and task adapters~\citep{adapterhub} but currently supports only about 50 languages, and hence most languages are not covered. More importantly, as mentioned in the introduction, certain languages have diverse regional varieties and difficulty of reliably obtaining data for them makes adapter-based approaches especially brittle in these cases. In the following \autoref{sec:generalized_adapters}, we propose strategies to improve the robustness of language adapters to uncovered languages without training new adapters.    

\section{Generalizing Language Adapters to Related Languages\label{sec:generalized_adapters}}
We consider the setting where we have a multilingual pretrained model $\mathcal{M}$ as well as the pretrained task adapters $\mathcal{T}_1,\mathcal{T}_2,...,\mathcal{T}_m$ and language adapters $\mathcal{L}_1,\mathcal{L}_2,...,\mathcal{L}_n$. We want to use $\mathcal{M}$ and the existing adapters to support a new language $L_\text{new}$, which is not in $\{L_1, L_2,...,L_n\}$ on a given task $T$ without training a new adapter for $\mathcal{L}_\text{new}$. 

\noindent \textbf{Related Language Adapters} One potential solution is to find the most related language $L_\text{rel} \in \{L_1,L_2,...,L_n\}$ and then use $\mathcal{T} \circ \mathcal{L}_\text{rel} \circ \mathcal{M}$ to do inference in $L_\text{new}$. However, this has two disadvantages. First, the task adapter $\mathcal{T}$ is only trained in the setting of $\mathcal{T} \circ \mathcal{L}_\text{src} \circ \mathcal{M}$, so it might not generalize well to the test time setting of  $\mathcal{T} \circ \mathcal{L}_\text{rel} \circ \mathcal{M}$~(as shown in \autoref{subsec:results}). Second, while the pretrained model $\mathcal{M}$ may be relatively robust against distribution shifts~\citep{hendrycks-etal-2020-pretrained}, the specialized language adapters might make the model brittle to language variations because they are trained for specific languages. Our experiments in \autoref{subsec:results} show that this solution indeed leads to poor performance. 

\paragraph{Adapter Ensembling}
As a first solution to this problem, we propose an extremely simple strategy of averaging the 
transformed outputs of multiple language adapters.
Specifically, we use both the source language adapter $\mathcal{L}_{\text{src}}$ and adapters from related languages with similar linguistic properties to the new language. Let $\mathcal{R}$ be the set of the source and related language adapters.
To do inference on a task $T$ for the new language $L_\text{new}$, we transform the output $h$ of each layer in $\mathcal{M}$ with the language adapters as 
$
   \mathcal{L}_\text{avg}(h) = \frac{1}{R} \sum_{i=1}^{R} \mathcal{L}_i(h)
$.

\paragraph{Entropy Minimized Ensemble of Adapters}
While ensembling is a simple and effective strategy to combine multiple potentially beneficial language adapters, the equal weighing of all language adapters could be sub-optimal for $L_\text{new}$; different language varieties, or even sentences, could benefit from a different weighting of the pretrained language adapters. To further improve adapter ensembling, we generalize $\mathcal{L}_\text{avg}(h)$ into a learnable weighted average:
$$
\mathcal{L}_\text{wavg}(h) = \sum\nolimits_{i=1}^{R} \alpha_i \mathcal{L}_i(h)
$$
where $\alpha_1, \alpha_2, ..., \alpha_R$ are learnable weights satisfying $\alpha_i \geq 0$ and $\sum_{i=1}^{S} \alpha_i = 1$. 
Next, we propose Entropy Minimized Ensemble of Adapters~(EMEA) method, which learns the adapter weightings \textit{for each sentence} without additional training.

The intuition behind our method is that a good adapter weight $\alpha$ for a test input $x$ should make the model more confident in its prediction for $x$, that is, it should lead to lower model entropy over the input~\citep{entropy,tent}.
Specifically for structured prediction tasks, we want to classify each word $x_w$ in a test input $x$ with $W$ words into one of the possible $C$ classes. 
We consider the entropy:
$
 \small
 H(x; \alpha)  = -\sum_{w=1}^{W} \sum_{c=1}^{C} P(c |x_w; \alpha) \log{P(c | x_w; \alpha)},   
$
where $P(c | x_w; \alpha)$ is the prediction of the model $\mathcal{T} \circ L_\text{wavg}(h) \circ \mathcal{M}$.
Since $P(c|x_w;\alpha)$ is a function of the ensemble weights $\alpha$, we can calculate the gradient of $\alpha$ as $g_i = \nabla_{\alpha_i} H(x; \alpha)$.

To minimize the entropy loss, we can simply do gradient descent steps on each $\alpha_i$ using the corresponding gradient $g_i$ by
$
\small
    \alpha_i = \alpha_i - \gamma g_i
$,
where $\gamma$ is the learning rate. We can then use the updated $\alpha$ to calculate the final prediction for $x$. In \autoref{sec:exp}, we find that a single step of gradient update already leads to better performance than simple ensembling. We can additionally perform multiple steps of gradient descent to obtain a better $\alpha$ at the cost of lower inference speed. \autoref{alg:alg} shows the pseudo code of our method\footnote{Code can be found at \url{https://github.com/cindyxinyiwang/emea}}.


\begin{algorithm}[t!]
\small
\SetAlgoLined
\DontPrintSemicolon
\SetKwInOut{Input}{Input}
\SetKwInOut{Output}{Output}
\SetCommentSty{itshape}
\SetKwComment{Comment}{$\triangleright$ }{}
\Input{Uniform weights $\alpha^0$, weighted adapter output; $L_\text{wavg}(h, \alpha^0)$; test data $x$; number of update steps $T$
}
\Output{Prediction $\hat{y}$}
   
  \For{t in 0, 1, ..., T-1}{
    
    \Comment{Calculate entropy}
    $H(x, \alpha) \leftarrow \text{Entropy}(\mathcal{T} \circ L_\text{wavg}(h, \alpha^t) \circ \mathcal{M})$
    
    \Comment{Calculate gradient}
    $g^t =  \nabla_{\alpha} H(x; \alpha^t)$
    
    \Comment{Update weighting}
    $\alpha^{t+1} \leftarrow \text{Update}(\alpha^t, g^t)$
    
  }
    \Comment{Calculate final prediction}
    $\hat{y} \leftarrow \text{Predict}(\mathcal{T} \circ L_\text{wavg}(h, \alpha^T) \circ \mathcal{M})$
\caption{ Training with EMEA}
\label{alg:alg}
\end{algorithm}
\section{Experiments\label{sec:exp}}
\begin{table}[t]
    \centering
     \resizebox{0.3\textwidth}{!}{%
    \begin{tabular}{c|l|l}
    \toprule
    Related  & Additional   &  Test \\
    \midrule
      hi & en,ar  & mr,bn,ta,bho   \\
      is & en,de & fo,no,da   \\
      ru & en & be,uk,bg   \\
    \bottomrule
    \end{tabular}}
    \caption{Test language groups and their corresponding language adapters. Adapters from languages in the first two columns are applied to the test languages in the third column.}
    \vspace{-1.5em}
    \label{tab:language}
\end{table}

\begin{table*}[]
    \centering
    \resizebox{0.75\textwidth}{!}{%
    \begin{tabular}{l|l|cccc|cccc|cccc|c}
    \toprule
     Task & Method &  mr & bn & ta & avg.  & fo & no & da & avg. & be & uk & bg & avg. & avg. \\
     \midrule
  \multirow{5}{*}{NER}  & En & 48.0 & 54.4 & 29.6 & 44.0 & 57.5 & 73.3 & 80.5 & 70.4 & 67.1 & 67.6 & 71.1 & 68.6 & 61.0 \\
    & Related & 51.7 & 47.0 & 30.8 & 43.1 & 54.3 & 72.7 & 79.3 & 68.7 & 66.2 & 65.8 & 69.8 & 67.3 & 59.7 \\
    & CL & 48.1 & 55.2  & 28.9 & 44.1 & 57.5 & 73.6 & 80.6 & 70.6 & 67.0 & 67.8 & 71.0 & 68.6 & 61.1 \\
    & Fusion & 49.8 & 58.3 & 33.7 & 47.2 & 56.0 & 69.3 & 77.8 & 67.7 & 70.1 & 69.1 & 72.3 & 70.5 & 61.8 \\
    \cmidrule{2-15}
    & Ensemble & 55.5 & 55.3 & 35.8 & 48.8 & 57.4 & 74.0 & 80.8 & 70.7 & 70.5 & 72.2 & 74.2 & 72.3 & 63.9 \\
    & EMEA-s1 & 57.2 & 61.2 & 37.4 & 51.9 & 59.2 & 74.3 & 81.3 & 71.6 & 71.5 & 72.9 & 74.9 & 73.1 & 65.5 \\
    & EMEA-s10 & \textbf{57.5} & \textbf{63.2} & \textbf{38.3} & \textbf{53.0} & \textbf{61.6} & \textbf{74.9} & \textbf{82.0} & \textbf{72.8} & \textbf{72.9} & \textbf{72.9} & \textbf{75.1} & \textbf{73.6} & \textbf{66.5} \\
     \bottomrule
     \toprule
    & Method  &  mr & bho & ta & avg. & fo & no & da & avg. & be & uk & bg & avg. & avg. \\
     \midrule
   \multirow{5}{*}{POS} & En  & 62.6 & 39.5 & 53.4 & 51.8 & 71.6 & \textbf{84.6} & 87.6 & 81.1 & 85.3 & 81.4 & 84.6 & 83.7 & 72.2 \\
   & Related  & 53.2 & \textbf{46.9} & 47.0 & 49.0 & 72.8 & 82.4 & 86.9 & 80.7 & 84.0 & 79.5 & 82.9 & 82.1 & 70.6 \\
   & CL  & 62.6 & 39.6 & 53.6 & 51.9 & 71.7 & 84.2 & 87.7 & 81.2 & 85.6 & 81.5 & 84.7 & 83.9 & 72.3 \\
   & Fusion  & 59.8 & 42.3 & 53.5 & 51.8 & 72.9 & 81.3 & 86.0 & 80.0 & 85.8 & 80.0 & 83.3 & 83.0 & 71.6 \\
    \cmidrule{2-15}
   & Ensemble  & 62.2 & 45.5 & 53.7 & 53.8 & 73.9 & 83.6 & 87.9 & 81.8 & 85.9 & \textbf{81.6} & 84.6 & 84.0 & 73.2 \\
   & EMEA-s1  & 62.1 & 45.1 & 54.3 & 53.8 & \textbf{74.0} & 83.5 & 87.8 & 81.7 & 86.2 & 81.4 & 84.6 & 84.0 & 73.2 \\
   & EMEA-s10 & \textbf{62.5} & 44.9 & \textbf{55.6} & \textbf{54.3} & 73.8 & 83.7 & \textbf{88.0} & \textbf{81.8} & \textbf{86.0} & \textbf{81.6} & \textbf{84.9} & \textbf{84.2} & \textbf{73.5} \\
    \bottomrule
    \end{tabular}}
     \vspace{-0.5em}
    \caption{F1 of the baselines and our methods for each language group. EMEA-s1 updates the adapter weights with a single gradient step while EMEA-s10 updates for 10 steps.}
    \label{tab:ner}
     \vspace{-1em}
\end{table*}

\begin{figure*}[t]
    \centering
    \begin{minipage}[t]{0.22\textwidth}
        \centering
        \includegraphics[width=0.98\linewidth]{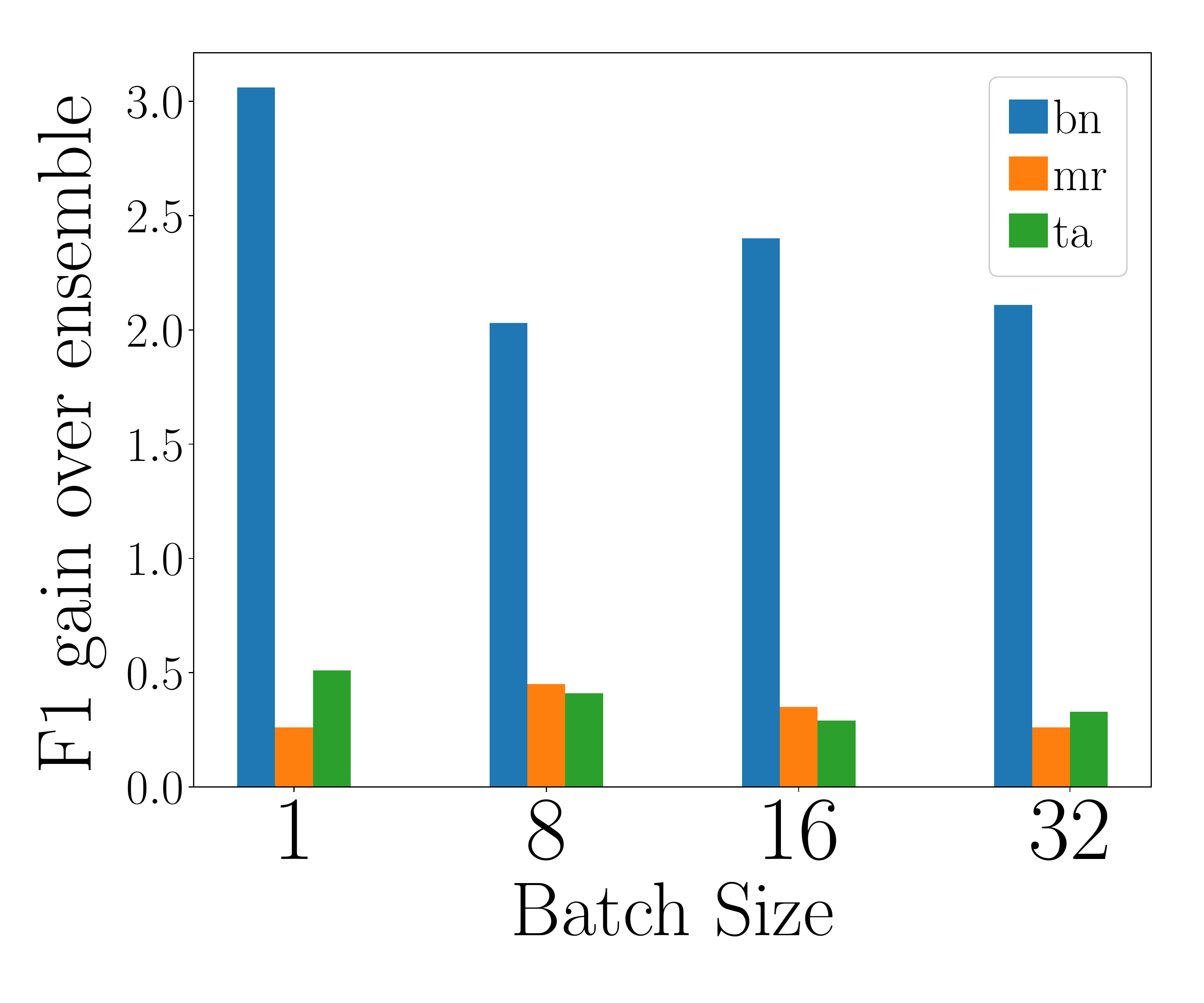}
        \vspace{-2em}
        \caption{Improvements over ensemble with different batch size.}
        \label{fig:bsize}
    \end{minipage}
    \hspace{5mm}
    \begin{minipage}[t]{0.23\textwidth}
        \centering
        \includegraphics[width=0.98\linewidth]{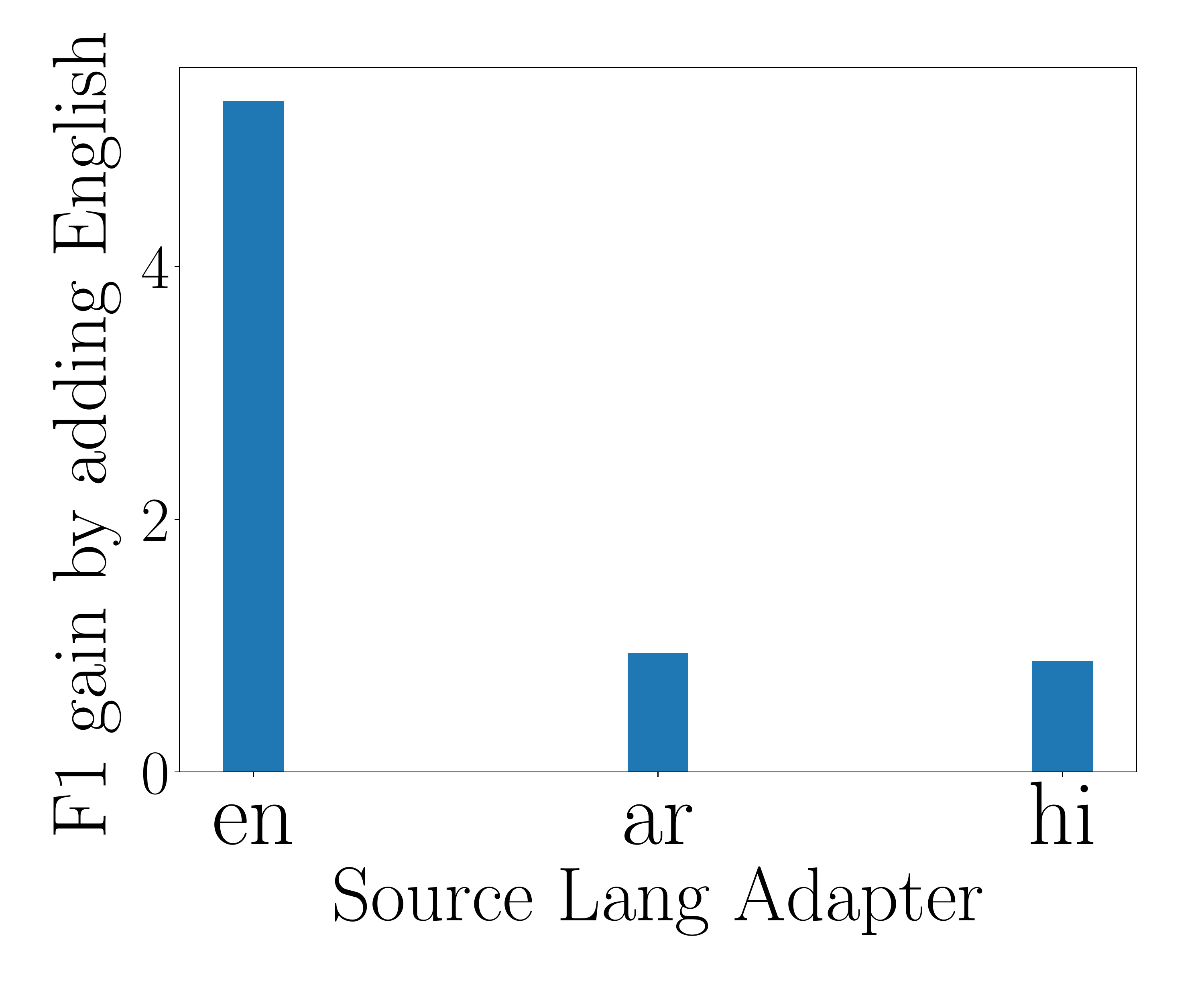}
        \vspace{-2em}
        \caption{Improvements by adding \texttt{en} adapter for different src language adapters.}
        \label{fig:src_lang}
    \end{minipage}
     \vspace{-0.5mm}
    \hspace{5mm}
        \begin{minipage}[t]{0.45\textwidth}
        \centering
        \includegraphics[width=0.98\linewidth]{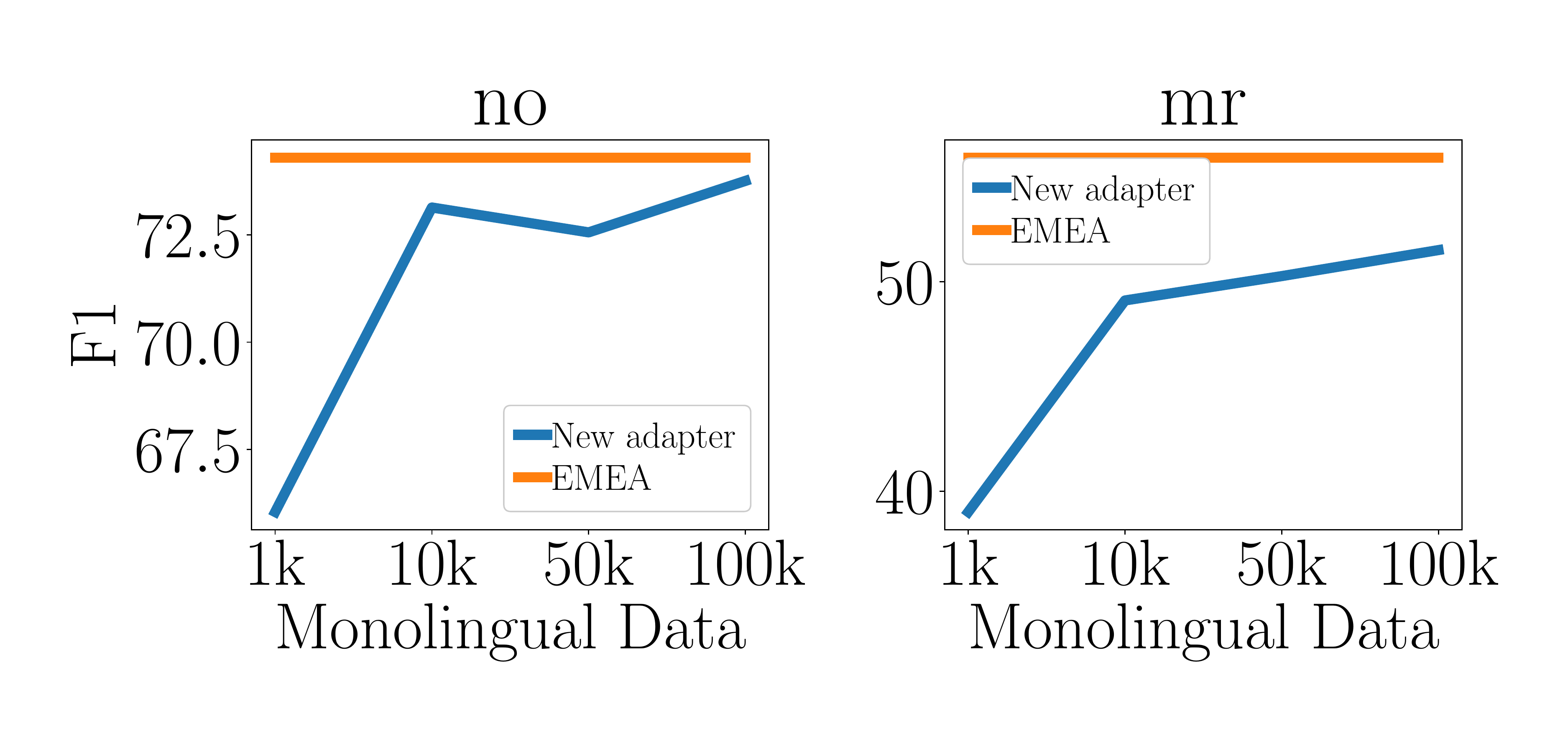}
        \vspace{-1em}
        \caption{Comparison to training adapter on different amount of monolingual data.}
        \label{fig:mono}
    \end{minipage}
     \vspace{-1.5em}
\end{figure*}


\paragraph{Data} We focus on zero-shot cross-lingual transfer with English as the source language. We conduct experiments on named entity recognition~(NER) and part-of-speech tagging~(POS). We use the WikiAnn dataset~\citep{Pan2017} for NER and Universial Treebank 2.0 for POS tagging~\citep{nivre2018universal}.

\paragraph{Model} We use the mBERT~\cite{bert} model, which shows good performance for low-resource languages on the structured prediction tasks~\citep{madx-pfeiffer-etal-2020,xtreme2020}. We use the English annotated data to train the task adapter. Each experiment is run with 3 different random seeds and we report the average performance. More details can be found in \autoref{app:train}.

\paragraph{Languages} Due to the lack of datasets for dialects, we focus on three groups of closely related languages to simulate the setup of language varieties. 
Each group has a language with a pretrained adapter available on the AdapterHub~\citep{adapterhub}, and we test on the languages without adapters. The language with adapter and the target languages for each group are: 1. Hindi~(\texttt{hi}): Marathi~(\texttt{mr}), Bengali~(\texttt{bn}), Tamil~(\texttt{ta}), Bhojpuri~(\texttt{bho}); 2. Icelandic~(\texttt{is}): Faroese~(\texttt{fo}), Norwegian~(\texttt{no}), Danish~(\texttt{da}); 3. Russian~(\texttt{ru}): Bulgarian~(\texttt{bg}), Ukrainian~(\texttt{uk}), Belorussian~(\texttt{be}). For our methods, we additionally use the adapter for English (the \emph{src} language), and optionally for another highly related language if there is one available on the AdapterHub. The adapters used are listed in \autoref{tab:language}.

\paragraph{Baselines} We compare with several baselines: 1) En: the English adapter; 2) Related: the best performing related language adapter; 
3) Continual learning (CL): we use the English language adapter and update its parameters using the entropy loss for each test input; 
4) Fusion: learn another set of key, value and query parameters in each layer that uses the layer output as a query to mix together the output of each adapter~\citep{adapterfusion}. Since we do not use labeled data in the new language, we train the fusion parameters on English labeled data. 

\subsection{Results\label{subsec:results}}
The results can be found in \autoref{tab:ner}. For most languages using the English adapter is better than the best individual related language adapter. This confirms our hypothesis that specialized language adapters are not robust to language variations. CL leads to slight improvements for some languages but is generally comparable to En.
Fusion improves over En for the NER task but it requires training and storing extra parameters. Its performance is also not consistent across languages and tasks, likely because it is only trained on English labeled data.

\paragraph{Using multiple language adapters brings significant gains} Ensembling leads to significant gains for the non-Latin language group. It also brings improvements or is comparable to the best baseline on other languages. EMEA delivers further improvements across almost all languages, demonstrating the effectiveness of adapting language adapter weights to each test sentence. With only a single gradient update step on the ensemble weights, EMEA-s1 already leads to significant improvements over ensembling for NER. EMEA-s10 brings additional improvements on both tasks because it learns more optimal ensembling weights with 10 gradient update steps~(we list the inference cost for each method in \autoref{app:decode_speed}). We hypothesize that the proposed methods improve non-Latin languages more because these are low-performing languages that the model is more uncertain about. 

\paragraph{Effect of test batch size}
In \autoref{fig:bsize} we plot the result of using different test batch sizes with EMEA on the NER task. A smaller batch size leads to more fine-grained test time adaptation with a higher computational cost. \autoref{fig:bsize} shows that a smaller batch size indeed leads to better performance while using a larger batch size still outperforms the baseline. 

\paragraph{Significance of source language adapter} We investigate whether the benefit of adding the \emph{src} language adapter comes from the discrepancy between training and testing of the task adapter. We train different task adapters with language adapters other than English~(\texttt{en}), and compare the improvement of adding the \texttt{en} adapter for the ensemble. \autoref{fig:src_lang} shows that the \texttt{en} adapter provides the largest benefit when it is used to train the task adapter, which verifies that using different language adapters with the task adapter between training and testing leads to sub-optimal cross-lingual transfer performance.

\paragraph{Comparison to training new adapters}
In order to better understand how much data is required to train new language adapters that are competitive with EMEA, we trained new adapters using a small amount of monolingual data in the target language. We focus on two languages, \texttt{mr} and \texttt{no}, on the NER task, and show the results in \autoref{fig:mono}. Note that this setting puts EMEA at a disadvantage because EMEA does not require any training. It takes about 100k monolingual data for \texttt{no} to reach comparable performance with our method, while \texttt{mr} still lags behind EMEA. As large amounts of monolingual data are difficult to obtain for many language varieties and under-represented languages, EMEA can serve as a useful baseline for applying NLP models to such low-resource settings.

\begin{figure}
    \centering
    \includegraphics[width=0.7\columnwidth]{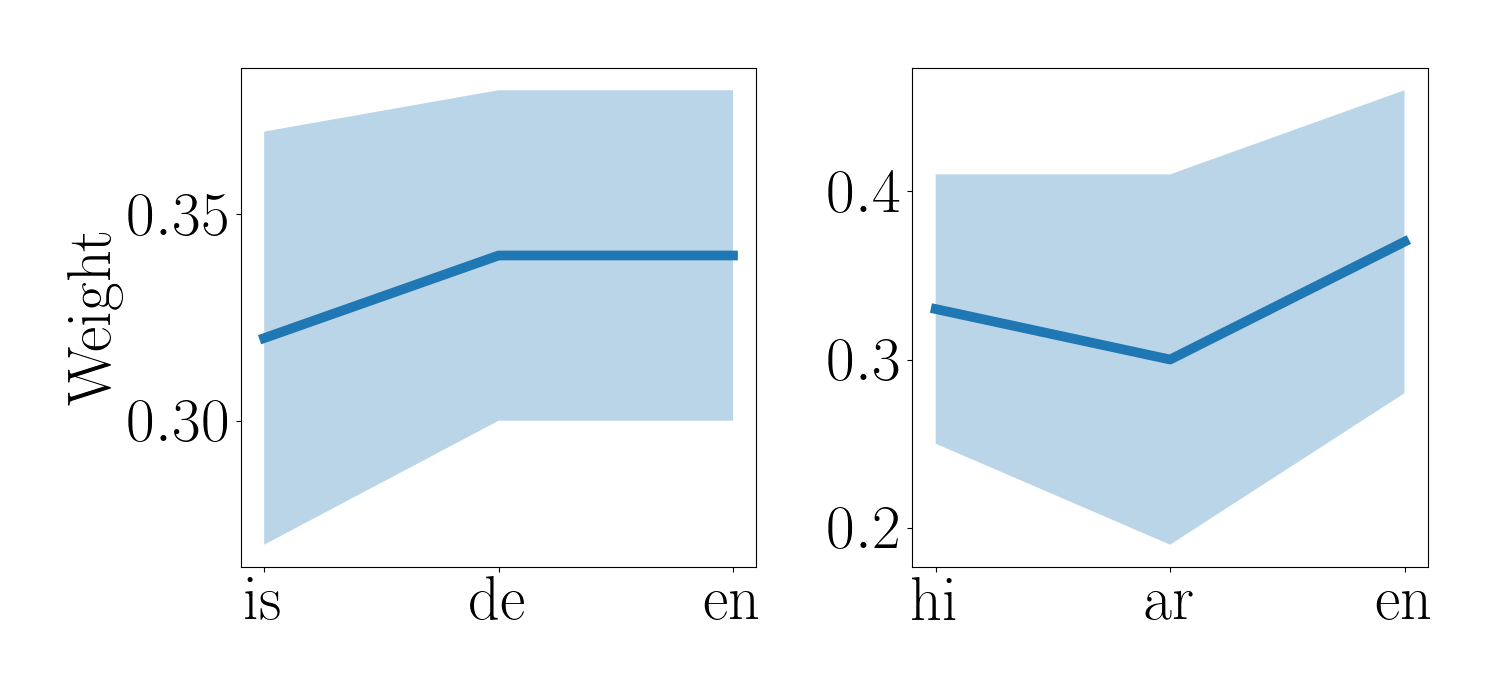}
    \vspace{-1em}
    \caption{Mean and standard deviation of the weight for each adapter for the \texttt{is}~(\textit{left}) and \texttt{hi}~(\textit{right}) language groups.}
    \vspace{-1.5em}
    \label{fig:weight}
\end{figure}

\paragraph{Analysis of weights} We plot the mean and standard deviation of ensembling weights from EMEA in \autoref{fig:weight}. The \texttt{En} adapter gets the highest weight for both language groups, in line with the results in \autoref{tab:ner} showing \texttt{en} as the best individual adapter. For the \texttt{hi} language group, the \texttt{ar} adapter tends to have the least benefit, probably because it has a different script from the languages we test on.  


\section{Related Work}
Our work is related to parameter efficient fine-tuning of pretrained models~\citep{adapter_nmt,madx-pfeiffer-etal-2020,prefixtuning,guo2020parameterefficient}. Specifically, \citep{udapter,karimi-mahabadi-etal-2021-parameter} make adapters more generalizable by learning a parameter generator, while our work aims to utilize existing pretrained adapters without further training. \citet{adapterfusion} propose to learn extra parameters using labeled data to combine pretrained multitask adapters whereas our method does not require any training or labeled data. While we focus on language adapters in this work, our method is also applicable to ensembling domain or task adapters. Finally, our method is inspired by the test time adaptation framework proposed for image classification~\citep{sun19ttt,tent,keep_learning}. Instead of adapting a single model, we focus on efficient utilization of many pre-trained language adapters to improve the model's robustness to language variations.  

\section{Discussion and Conclusion}
Language and dialect cannot be simply categorized into monolithic entities. Thus a truly intelligent NLP system should be able to recognize and adapt to personalized language varieties after it is trained and deployed. However, the standard system evaluation is built on the assumption that an NLP model is fixed once it is trained. In this paper, we focus on a specific case of this general problem---we find that specialized language adapters might not be robust to unseen language variations, and that utilization of multiple existing pretrained language adapters alleviates this issue. We hope our findings can inspire future work on models that are robust and adaptive to language variations.

We identify two limitations of this paper, which we leave to future work. First, there are limited datasets and benchmarks that evaluate NLP models' ability to generalize to unseen dialect variations. Therefore, we only test our method on NER and POS tagging tasks because they have the best language coverage. It is an important future direction to construct high-quality datasets that consider language and dialect variations. Second, our method has slower inference speed due to test time computation. Future work can aim to reduce the cost by algorithmic or hardware innovations.  

\section*{Acknowledgement}
This material is based on work supported by the National Science Foundation under grants 2040926 and 2125201. XW is supported by the Apple PhD fellowship. The authors would like to thank Laura Rimell, Sachin Kumar and Hieu Pham for helpful discussions on the drafts of the paper.
\bibliography{anthology,custom}
\bibliographystyle{acl_natbib}

\clearpage
\appendix
\section{Implementation Details\label{app:train}}
 We preprocess the data using scripts in XTREME~\citep{xtreme2020}. We use the best performing adapter configuration in \citet{madx-pfeiffer-etal-2020}. For NER, we train the task adapter for 100 epochs using learning rate of 1e-4. For POS tagging, we train the task adapter for 50 epochs with the learning rate of 1e-4. For EMEA, we search over the learning rate $\gamma$ of ${0.1, 1, 10}$ on the English validation set and pick $\gamma=10$ for all experiments.

For Fusion, we use learning rate of 5e-5 which is recommended by \citep{adapterfusion}. We search over the best learning rate for CL on the performance of English labeled data. We use the learning rate of 2e-5 and do 1 step of gradient update for each batch. 

For our experiment on training new adapters, we find that training from scratch on \texttt{no} and \texttt{mr} is not competitive when using very small amount of data. Therefore, we continue training from their related language adapters.

\section{Decoding Speed\label{app:decode_speed}}
We list the inference time for various methods in the paper in \autoref{tab:decode_speed}. EMEA leads to better performance at a cost of lower inference speed. We leave it to future work to explore strategies that speed up the test time optimization.
\begin{table}[t]
    \centering
    \resizebox{0.3\textwidth}{!}{%
    \begin{tabular}{l|l}
    \toprule 
      Method  & Example/Second  \\
      \midrule 
       Single Adapter  & 250 \\
       CL & 77 \\
       Fusion & 200 \\
       Ensemble & 200 \\
       EMEA-s1 & 62 \\
       EMEA-s10 & 9 \\
    \bottomrule
    \end{tabular}}
    \caption{Decoding speed for different methods used in the paper.}
    \label{tab:decode_speed}
\end{table}

\section{Examples of outputs}
We compare the outputs of EMEA with the best baseline on the POS tagging task for Norwegian~(\texttt{no}). Although both methods struggle with verb and adjective predictions, EMEA is often better at predicting the correct adjectives compared to the baseline.

\begin{table}[]
    \centering
      \resizebox{0.4\textwidth}{!}{%
    \begin{tabular}{l|l}
    \toprule
       src  &  Lendið, er, kargt, og, oyði, . \\
       tgt & NOUN, VERB, ADJ, CCONJ, ADJ, PUNCT  \\
       Base & NOUN, \textcolor{red}{AUX}, ADJ, CCONJ, \textcolor{red}{NOUN}, PUNCT \\
       EMEA & NOUN, \textcolor{red}{AUX}, ADJ, CCONJ, \textcolor{blue}{ADJ}, PUNCT \\
    \midrule
    src & Útvinningin, er, í, tveimum, umførum, . \\
    tgt & NOUN, VERB, ADP, NUM, NOUN, PUNCT \\
    Base & NOUN, \textcolor{red}{AUX}, ADP, \textcolor{red}{ADJ}, NOUN, PUNCT \\
    EMEA & NOUN, \textcolor{blue}{VERB}, ADP,  \textcolor{blue}{NUM}, NOUN, PUNCT \\
    \bottomrule
    \end{tabular}}
    \caption{Example outputs on POS tagging.}
    \label{tab:my_label}
\end{table}

%

\end{document}